\pdfoutput=1

\documentclass[11pt]{article}

\usepackage[preprint]{acl}
\usepackage{stfloats}
\fnbelowfloat 

\usepackage{amssymb,graphicx,color}
\usepackage{amsfonts}
\usepackage{latexsym}
\usepackage{float}
\usepackage{amsmath}

\usepackage{pdfpages} 

\definecolor{Red}{rgb}{1,0.5,0.5} 
\definecolor{Green}{rgb}{0.5,1,0.5} 
\definecolor{Blue}{rgb}{0.5,0.5,1} 
\usepackage{times}
\usepackage{latexsym}

\usepackage[T1]{fontenc}

\usepackage[utf8]{inputenc}

\usepackage{microtype}
\usepackage{hyperref}
\usepackage{listings}
\usepackage{xcolor}
\usepackage{amssymb,graphicx,color}
\usepackage{amsfonts}
\usepackage{latexsym}
\usepackage{float}
\usepackage{amsmath}

\usepackage{inconsolata}

\usepackage{graphicx}

\title{
    Quantifying Generative Media Bias with a Corpus of Real-world and Generated News Articles
}

\author{Filip Trhlik \and Pontus Stenetorp \\
  University College London\\
  \texttt{filip.trhlik.21@ucl.ac.uk} \\ \texttt{p.stenetorp@cs.ucl.ac.uk}}

\begin{document}
\maketitle
\begin{abstract}
Large language models (LLMs) are increasingly being utilised across a range of tasks and domains, with a burgeoning interest in their application within the field of journalism. This trend raises concerns due to our limited understanding of LLM behaviour in this domain, especially with respect to political bias. Existing studies predominantly focus on LLMs undertaking political questionnaires, which offers only limited insights into their biases and operational nuances. To address this gap, our study establishes a new curated dataset that contains 2,100 human-written articles and utilises their descriptions to generate 56,700 synthetic articles using nine LLMs. This enables us to analyse shifts in properties between human-authored and machine-generated articles, with this study focusing on political bias, detecting it using both supervised models and LLMs. Our findings reveal significant disparities between base and instruction-tuned LLMs, with instruction-tuned models exhibiting consistent political bias. Furthermore, we are able to study how LLMs behave as classifiers, observing their display of political bias even in this role. Overall, for the first time within the journalistic domain, this study outlines a framework and provides a structured dataset for quantifiable experiments, serving as a foundation for further research into LLM political bias and its implications.

\end{abstract}

\section{Introduction}
The current generation of LLMs has emerged as an important factor in the ongoing digital transformation~\citep{dellacqua2023navigating}. ChatGPT, in particular, has become the fastest adopted technology as of 2024~\citep{humlum2024adoption}. These models have shown a clear impact across various fields, such as software development~\citep{Russo2023NavigatingTC} and academia~\citep{Fecher2023FriendOF}. 

The specific sector we aim to focus on is journalism. It has, in recent years, experienced a transformative period~\citep{martens2018digital}, moving from printed to digital news and grappling with the increased importance of social media and a subsequent rise of disinformation~\citep{Guess_Lyons_2020}. Recent reports~\citep{newman2023a} suggest that generative AI will bring the next significant shift, with 28\% of publishers reportedly using AI in their processes in 2023, and OpenAI expressing interest in this field.\footnote{\url{https://openai.com/blog/openai-and-journalism}}

The integration of LLMs into journalism promises new avenues for content creation and dissemination~\citep{nishal2024envisioning}. However, it comes with a set of challenges and considerations, which are particularly pressing given the importance of quality journalism to the functioning of a free and democratic society~\citep{christians2010normative}. LLMs notably face a multitude of problems, from style alignment~\citep{Shanahan2023EvaluatingLL} to bias management~\citep{Gallegos2023BiasAF}. Nevertheless, when it comes to utilising LLMs in journalistic processes, one of the most critical yet limitedly understood text properties is political bias.

Unchecked, it can significantly impact how people consume information—even with regard to verifiable facts—and form opinions on them~\citep{10.1257/pandp.20201072}. Its presence in media has been shown to enhance polarisation in society and a rise in extremism~\citep{10.1162/qjec.122.3.1187}. Therefore, this study seeks to investigate the occurrence of political bias directly in generated news articles, an aspect not covered in existing literature. By examining the extent to which current LLMs exhibit political bias in their generated content, in what direction this bias leans, and under what conditions it manifests, we can gain a concrete understanding of this issue and better assess the risks it poses. To answer these questions, this work delivers:
\begin{enumerate}
    \item A dataset for the comparative evaluation of generated and human-written news articles
    \item An analysis of political bias within nine LLMs, detecting the political bias using both supervised models and LLMs
    \item An assessment of political bias exhibited by LLMs classifying political leaning, showcasing how differently LLMs perceive their own outputs compared to other texts
\end{enumerate}
\section{Related Work}
\subsection{Political Bias Assessment}
Political bias refers to a predisposition towards a specific political ideology, party, or policy. It is a phenomenon that can significantly influence the presentation and reception of information in various forms of communication, including news articles, opinion pieces, blogs, and social media content~\citep{RePEc:plo:pone00:0193765}.

When evaluating political leaning, it is essential to specify which facet of bias is being examined. The two most common scales are economic (left-right)~\citep{Gold_1998} and social (authoritarian-libertarian)~\citep{Lane_1956}. Notably, the majority of current AI assessments of political bias focus on the economic scale.

Detecting political bias remains a challenging endeavour, with numerous recent publications aiming to refine the approach to this task. Within the sentence-level approach, some notable datasets include the \textit{Ideological Books Corpus}~\citep{iyyer-etal-2014-political}, \textit{BASIL}~\citep{chen-etal-2020-analyzing}, and \textit{BABE}~\citep{spinde-etal-2021-neural-media}. However, despite the progress in dataset development, in comparison to other detection tasks, the availability of data for sentence-level political bias detection remains limited.

In article-level political bias detection, a significant body of work focuses on addressing the same challenge through the use of external information~\citep{feng2021kgap, zhang-etal-2022-kcd}; however, this cannot be used for the purposes of this work. Additionally, the majority of recent work relies on data released by \textit{AllSides.com}, which categorises sources into five classes (left, left-leaning, centre, right-leaning, right). Various approaches have been tested with this data, such as a multi-view document attention model that utilises the title, link structure, and article content~\citep{kulkarni-etal-2018-multi}. The most notable work using \textit{AllSides.com} is the dataset by~\citet{baly-etal-2020-detect}, which labels the data per article, providing a more precise training set. Beyond the English language, the dataset by \citet{TerAkopyan2021} includes German news sources, offering data that represents perspectives outside the Anglo-American political sphere.

Finally, on the topic of detection of political bias,~\citet{liu-etal-2022-politics} proposed a novel improvement involving pretraining a RoBERTa-base language model on 3.6 million news articles with ideology-driven pretraining objectives. This model, \textit{POLITICS}, currently achieves SOTA results across various benchmarks when fine-tuned appropriately, outperforming RoBERTa by up to 10\% in accuracy.

\subsection{Classification Through LLMs}
Due to the limitations of trained classification models, which often suffer from performance issues when confronted with out-of-domain data not included in the training datasets, the use of LLMs as classifiers has been proposed~\citep{lin-etal-2024-indivec}. Studies have demonstrated that LLMs can handle simple classification tasks and even achieve SOTA results~\citep{sun2023text}. However, using LLMs for political bias detection is more complex, as they exhibit biases in their classifications~\citep{lin2024investigating}. Despite this, LLMs offer a novel evaluation method that allows for further bias assessment, without the constraints of limited training data, and that provides deeper insights into the biases of the LLMs used for classification.

\subsection{Political Bias in LLMs}
\label{Political Bias in LLMs}
LLMs are prone to various types of bias~\citep{ganguli2023capacity}. Here, we focus specifically on their political bias, a relatively recent topic that has gained more attention following the popularisation of GPT-3.5. Current studies have adopted various methodologies to measure it.~\citet{urman_makhortykh_2023} measured bias by monitoring which questions LLMs refuse to answer.~\citet{Motoki2023} explored the responses GPT-3.5 provided to questions from the Political Compass questionnaire under various prompts. The most recent study, by~\citet{feng-etal-2023-pretraining}, also used the Political Compass but examined a broader range of models to understand how biases impact performance. They found that the political bias of an LLM influences the fairness of downstream NLP models trained on top of it, underscoring the significance of bias in these models. The main limitation of all these works is their reliance on the questionnaires and other methods that are detached from the biases directly evident in the content generated by LLMs. Thus, these methods cannot be directly equated to LLMs exhibiting the same biases in generating journalistic text.

\section{Task \& Data}
\subsection{Task Specification}
As noted, current studies typically use political questionnaires to examine political bias in LLMs. Since LLMs will likely be used in journalism to generate articles, we see it as imperative to assess political bias directly in the generated text rather than through proxy tests. However, this approach presents several challenges. Firstly, it requires generating numerous texts that can be compared in terms of political bias while also maintaining a journalistic nature. Secondly, from an assessment standpoint, the bias of the classifiers themselves must be addressed~\citep{ntoutsi2020}. Classifying only the generated articles can yield very different results depending on the classifiers and their corresponding training sets~\citep{VANGIFFEN202293}.

To address these challenges, we propose establishing a new dataset that utilises news articles and their summaries authored by journalists to generate articles from the perspective of LLMs, using the summaries as part of the generative prompts. This approach can resolve the aforementioned issues, as the human-authored articles will serve as anchor points, allowing us to measure only the relative shift in bias between the real article and the one generated from its summary. Since all models will use the same set of summaries and their outputs will be compared against the same set of human-written articles, the assessment quantification issue should be mitigated.

Among all news summary datasets, the \textit{NEWSROOM} dataset by~\citet{grusky-etal-2018-newsroom} emerges as the most suitable for us to obtain data of sufficient quality. It offers 1.3 million complete articles covering a wide array of topics from 38 distinct sources. Furthermore, the accompanying summaries are written by journalists and editors, providing enough relevant information about the articles to satisfy our criteria.

\subsection{Categorisation of News Articles}
To filter our data and enhance informativeness, we fine-tune two models for classifying news categories. We use \textit{20 Newsgroup} dataset by~\citet{Lang95}, which divides articles into 8 categories with 20 subcategories, and \textit{News Category Dataset} by~\citet{misra2022news} that offers a more granular categorisation across 42 classes.

For our experiment, we want to select articles from categories likely to elicit political disagreement. We choose the topics of \textbf{Politics} and \textbf{Business} due to their clear potential for socio-economic political biases. Additionally, we include \textbf{Sport} as a control sample, given its usual detachment from political discourse. Finally, we select topics such as \textbf{Religion}~\citep{doi:10.1177/2050303213506476}, \textbf{LGBT}~\citep{Nolan2019}, \textbf{Ecology}~\citep{10.3389/fevo.2017.00175}, and \textbf{Guns}~\citep{Jashinsky2016} for their relevance and capacity to incite diverse opinions. As the two training datasets cover a different set of categories, we utilise them both to comprehensively categorise the news data. Appendix~\ref{tab:category_assignment} illustrates how our classes are linked to the classes from the training datasets. When both datasets feature a relevant class, we use them concurrently to define the new category in our training data. If only one dataset includes the relevant class, we disregard the label from the model trained on the other dataset.

With the categories selected, RoBERTa models~\citep{liu2019roberta} are fine-tuned for classification with hyperparameters detailed in Appendix~\ref{table:hyperparameters}. The two trained models label each article in the \textit{NEWSROOM} dataset. Articles that do not receive any label combination of interest are discarded. In Appendix~\ref{fig:WC_groups}, the most frequent unique words per category are presented, helping us confirm accurate categorisation of the articles.

\subsection{Data Selection}
After categorisation, we need to select an ideal subset of the remaining data for our new dataset. As such, we can discard any data that is not optimal for subsequent generation or comparison steps. Selection will focus on three main criteria: the \textbf{article length}, the \textbf{summary length}, and the \textbf{summary metrics} (compression, coverage, density).

\paragraph{Article Length}
In both generation and classification tasks, our objective is to utilise a maximum input/output size of 512 tokens. For current transformer-based models, one token approximately corresponds to four characters in the English language~\citep{sennrich-etal-2016-neural}. To ensure articles are sufficiently representative, we aim to evaluate at least 25\% of each, selecting only articles \textbf{between 1,000 and 8,000 characters} in length.

\paragraph{Summary Length}
Given the limited context window of generative LLMs, we exclude articles with summaries that \textbf{exceed 500 characters}.

\paragraph{Summary Metrics}
Lastly, we use the summary metrics from the \textit{NEWSROOM} dataset to exclude summaries that are unsuitable. \textbf{High compression} summaries, which may not adequately represent the full article, are discarded. Similarly, we discard summaries labelled with \textbf{low coverage} and \textbf{abstractive density} as they deviate significantly from describing the precise topic of their article.
\\\\
Finally, we select 300 articles from each category by sorting the remaining ones based on the length of their summaries, choosing those with the longest summaries in each category. This method results in a refined dataset comprising 2,100 articles and their summaries that are well-suited for generating synthetic articles.

\section{Article Generation}
Our aim is to generate synthetic news articles related to human-authored articles using their summaries in the prompts. To broaden our experiment, we will not only explore generation under unbiased conditions but also investigate deliberately biased settings. In these, models are prompted to emulate either a left-wing or a right-wing news style.

\subsection{Prompts}
Table~\ref{tab:generation_prompts} presents the prompt templates for both unbiased and biased settings. Our prompts are devised in a zero-shot manner to prevent any inadvertent introduction of bias. Additionally, the prompts specify the output length to align the average response length with our desired target length.\footnote{\url{https://platform.openai.com/docs/guides/prompt-engineering}}

We categorise the prompts according to two criteria: bias setting and prompt class. The bias setting distinguishes among three types of bias: left-wing, right-wing, and unbiased, steering the generation towards a specific political bias without referencing any particular news source. Additionally, the prompts vary by their prompt class to accommodate differently trained LLMs~\citep{Ouyang2022TrainingLM}. They can be either continuous for base LLMs or instructional for instruction-tuned LLMs.

Finally, we conducted a manual review to identify errors in the generations and to pinpoint any flaws in the models when confronted with specific prompts. The prompts listed in Table~\ref{tab:generation_prompts} represent the final, refined versions. When utilised, the \texttt{\{summary\}} placeholder in the prompts is replaced with the actual summary pertaining to the article being synthetically replicated. Thus, each human-written article corresponds to three different synthetic articles per model.

\begin{table*}[h!]
    \centering
    \begin{tabular}{l|p{6.8cm}|p{5.5cm}} 
    \hline
         & \textbf{Instruction} & \textbf{Continuous} \\ 
         \hline
         \textbf{Left-wing} & Write a 250 words long body of a left-wing news article that has a summary of "\{summary\}". Do not repeat the summary. & The following is the 250-word full text of a left-wing news article that has a summary of "\{summary\}":\\ 
         \hline
         \textbf{Unbiased} & Write a 250 words long body of a news article that has a summary of "\{summary\}". Do not repeat the summary. & The following is the 250-word full text of a news article that has a summary of "\{summary\}":\\ 
         \hline
         \textbf{Right-win} & Write a 250 words long body of a right-wing news article that has a summary of "\{summary\}". Do not repeat the summary. & The following is the 250-word full text of a right-wing news article that has a summary of "\{summary\}":\\
         \hline
    \end{tabular}
    \caption{Generation prompts}
    \label{tab:generation_prompts}
\end{table*}
\subsection{Model Selection}
Our aim is to incorporate an array of models with varying parameter sizes and from different sources (open or closed). Contemporary relevant models are listed in Table~\ref{tab:generative_models}, with third-party fine-tuned versions of the listed models excluded to allow more focus on the original LLMs. Many of these LLMs also exist in multiple versions; in such cases, we select the largest model that can be operated on an A100 80GB GPU. This approach follows the established precedent that larger models yield better text quality~\citep{Sarvazyan2023}.

\subsection{Generation Settings}
Our experiment assumes that the author of the generated articles seeks to attain the highest possible text quality. Therefore, we have chosen the settings that enhance text quality to reflect this target.

Regarding general settings, we set the output length at 512 tokens. This limit ensures computational efficiency by avoiding the generation of unnecessary tokens, considering that all our text evaluation methods have a maximum context length of 512 tokens, making generation of additional text redundant. Moreover, we opted for 32-bit float precision for open-source models to avoid any degradation of text quality.

In terms of decoding strategy, we selected the sampling strategy for our experiment. This choice is motivated by its reported superior performance regarding text quality~\citep{wiher2022decoding} and its ability to avoid degenerative repetition~\citep{holtzman2019curious}. Specific details regarding the decoding settings are available in Appendix~\ref{table:hyperparameters_gen}.

Lastly, we implemented a repetition penalty for GPT-2, as its use is recommended for this model; during testing, we observed that its outputs significantly benefited from this penalty~\citep{keskar2019ctrl}. Conversely, the larger models did not employ this penalty, as they performed effectively without it and its application appeared to directly harm the quality of some.
\begin{table*}[h]
\centering 
\begin{tabular}{l|p{3.8cm}|p{1.7cm}|p{1.7cm}} 
\hline
\textbf{Model}&\textbf{Identifier}&\textbf{Context window}&\textbf{Prompt type}\\
\hline
\textbf{GPT-2}~\citep{Radford2019LanguageMA}&gpt2-xl&1,024 &Continuous\\
\textbf{GPT-3.5}~\citep{chatgpt-etal-2022}&gpt-35-turbo-0613&4,096  & Instruction\\
\textbf{GPT-4}~\citep{openai2023gpt4}&gpt4-1106& 128,000 & Instruction\\
\textbf{Gemma}~\citep{gemma_2024} &gemma-7b& 8,192 & Continuous\\
\textbf{Gemma Instruct}~\citep{gemma_2024} &gemma-7b-it& 8,192& Instruction\\
\textbf{Mistral}~\citep{jiang2023mistral}&Mistral-7B-v0.1& 8,192& Continuous\\
\textbf{Mistral Chat}~\citep{jiang2023mistral}&Mistral-7B-Instruct-v0.1&8,192 & Instruction\\
\textbf{Llama 2}~\citep{touvron2023llama}&Llama-2-13b-hf&  4,096 & Continuous\\
\textbf{Llama 2-Chat}~\citep{touvron2023llama}&lama-2-13b-chat-hf&  4,096 & Instruction\\
\hline
\end{tabular}
\caption{Information regarding the LLMs used in our experiment}
\label{tab:generative_models}
\end{table*}
\subsection{Generation Process}
A total of 56,700 news articles were generated using nine different models. The main challenge encountered was the degenerative outputs. These included the insertion of symbols not part of the Latin alphabet, such as Arabic and Cyrillic characters, and HTML fragments. Smaller models, in particular, showed a tendency to produce such errors. To address this, we implemented a filter that regenerates the output whenever this behaviour occurs.

Moreover, we had to ensure that the generated articles were sufficiently long, especially with GPT-2 and Gemma Instruct often producing shorter texts. To minimise inconsistencies, we regenerated all texts under 500 characters until they reached a satisfactory length. The average length of a synthetic article in our dataset is 1,808 characters.

Similarly, we used the Self-BLEU metric~\citep{zhu2018texygen}, which measures the repetitiveness of generations, for spotting faulty ones. Lower scores indicate higher diversity, reflecting a broader range of sentence structures and more varied content within the articles. We marked any generation with a Self-BLEU score greater than 30 for regeneration, aiming to mimic the distribution of Self-BLEU scores found in human-written articles. The figure in Appendix~\ref{fig:bleu} illustrates the final Self-BLEU results in our dataset. While these measures do not completely eliminate all flaws, they ensure that the dataset adheres to basic quality standards.

At the end, we also regenerated all outputs containing warning messages and removed any mentions of political bias labels that persisted in the generated articles from the prompts.

\section{Experiment}
With all the data generated, the final dataset consists of 2,100 human-written articles and 56,700 AI-generated articles, each AI-generated article corresponding to a human-written counterpart.

The primary aim of the entire experiment is to investigate the political bias of LLMs quantitatively. This involves measuring the \textit{political bias shift}. To calculate this, we first determine the political alignment score for each article. This score is computed based on the probabilities that an article exhibits right-wing ($P_{\text{right}}$) or left-wing ($P_{\text{left}}$) tendencies. The formula for the political alignment score is:
\begin{equation}
\label{val}
\mathit{PA}_{\text{article}} = P_{\text{right}} - P_{\text{left}}
\end{equation}
We then calculate the alignment scores for both the AI-generated ($\mathit{PA}_{\text{generated}}$) and human-written ($\mathit{PA}_{\text{human}}$) articles. With human-written articles acting as a baseline, the political bias shift ($\Delta \mathit{PA}$) is calculated as follows:
\begin{equation}
\label{delta}
\Delta \mathit{PA} = \mathit{PA}_{\text{generated}} - \mathit{PA}_{\text{human}}
\end{equation}
The resulting value of $\Delta \mathit{PA}$ reflects the change in political alignment between the human-written and synthetic article. A positive $\Delta \mathit{PA}$ indicates a shift towards a right-wing bias in the AI-generated article, while a negative $\Delta \mathit{PA}$ suggests a left-wing bias. Overall, this approach enables us to study biases within LLMs by focusing solely on the shifts, allowing for comparative and quantitative analysis.

\subsection{Deep Learning Assessment}
With the metric established, the core experiment lies within measuring this political bias shift for all LLMs. Given the persistent concerns about bias in the evaluators, this study employs two distinct evaluation approaches. The first one involves a standard supervised model for bias detection.

Our training set includes the \textit{AllSides} dataset by~\citet{baly-etal-2020-detect} and the German News dataset by~\citet{TerAkopyan2021}, aiming to create a balanced dataset not skewed by exclusively US-centric data. We used the Google Translate API to convert German texts into English, manually verifying the translation quality. Although this translation might diminish stylistic diversity, the inclusion of translated data should still be advantageous ~\citep{10.1162/tacl_a_00593}. To mitigate overfitting and enhance model performance, we trained both the RoBERTa-large and POLITICS language models. Notably, in the test samples, the POLITICS model demonstrated a bias towards left-leaning classifications, while RoBERTa exhibited a tendency towards right-leaning ones. The averaging of their results yielded the least bias. We therefore adopted this practice for calculating political bias in this study.

\subsubsection*{Political Bias Shift Evaluation}
With the models fine-tuned, we classify all human-authored and generated articles, obtaining political shifts through equations~\ref{val} and~\ref{delta}. The results are shown in Figure~\ref{fig:political-shift}, with more detailed ones listed in Appendix~\ref{fig:PBS_large}. The figures list the average political shifts for all models studied across three prompt type settings: left-wing, unbiased, and right-wing.

\begin{figure*}[h]
    \centering
    \includegraphics[width=0.98\textwidth]{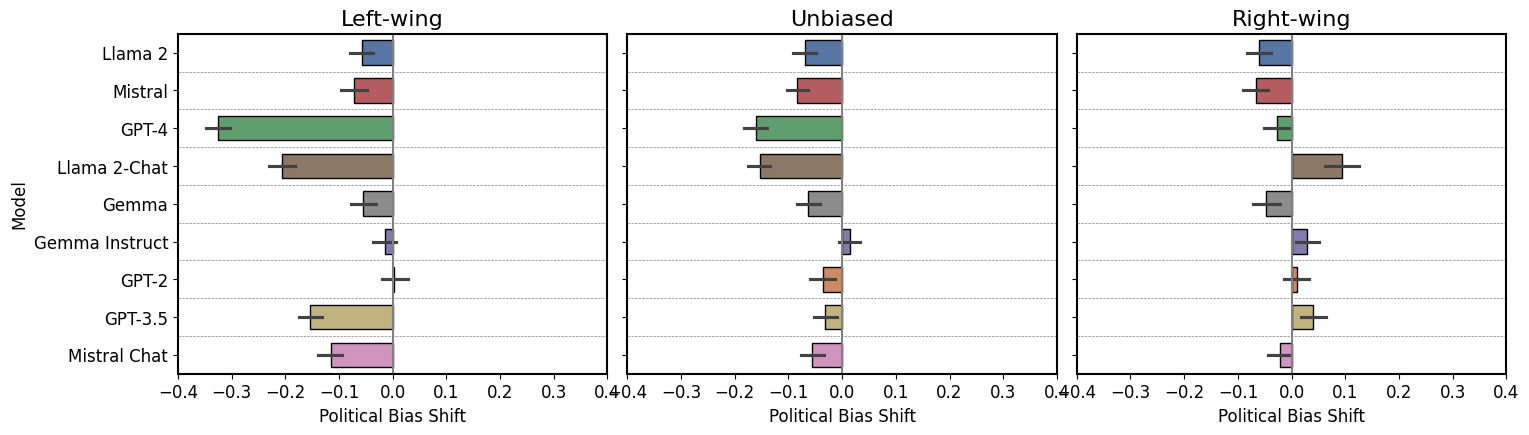}
    \caption{Political shift per model and prompt type (as assessed by supervised models)}
    \label{fig:political-shift}
\end{figure*}

\begin{figure*}[h]
    \centering
    \includegraphics[width=0.98\textwidth]{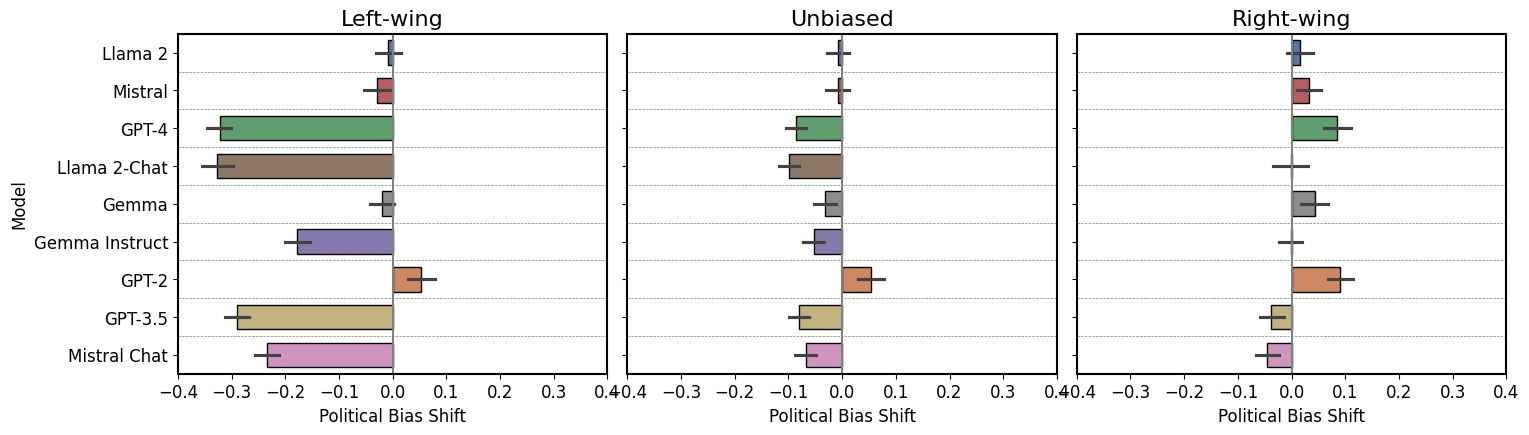}
    \caption{Political shift per model and prompt type (as assessed by LLMs) }
    \label{fig:political-shift-LLM}
\end{figure*}
The most notable trend is that the average political bias of the models skews left, regardless of the prompt types or news categories. This observation aligns with the literature in Section~\ref{Political Bias in LLMs}.

Moreover, the distinct prompt types are particularly noteworthy in this experiment, as they correlate directly with political bias. For all the instruction-tuned models, except Gemma Instruct, the desired bias is evident in the output data: \textbf{left news} prompts yield the most left-wing bias, and \textbf{right news} prompts yield the most right-wing bias. This indicates that these models possess an understanding of political biases. Notably, the \textbf{unbiased} prompts still tend towards the left.

Two models with notable behaviour for instruction-tuned LLMs are Gemma Instruct and Llama 2-Chat. Gemma Instruct predominantly produces unbiased content, irrespective of the prompt type. Conversely, Llama 2-Chat mostly aligns with the other instruction-tuned LLMs but exhibits a right-wing bias when prompted to generate it.

This alignment with prompt types does not hold for base models, which show no consistent response to the specifics of the prompts. Their outputs slightly lean towards the left but still occupy a more centric position compared to other models.

Finally, we note that GPT-4 emerges as the most politically biased model with a strong left-wing bias, followed by GPT-3.5 and Llama 2-Chat.
\subsection{LLM Assessment}
With the bias of the classifiers still a concern, our second evaluation approach aims to provide further data to ground our results and provide further insight into the political bias of various LLMs. This strategy employs the LLMs themselves to assess the political bias of news texts. To optimise the accuracy of LLM classifications, we employ the CARP technique by~\citet{sun2023text}, which defines a well-performing classification-prompt structure. The format of these prompts is detailed in Appendix~\ref{tab:CARP}. For the de-biased test set from the \textit{AllSides} dataset, employing GPT-3.5 as a classifier achieved an accuracy of 59.8\%, surpassing that of supervised models trained on this dataset~\citep{baly-etal-2020-detect}. This method proves sufficiently informative while unaffected by any specific dataset bias, thus making it particularly valuable for confirming results derived from supervised models. Moreover, it facilitates an exploration of how different LLMs manage classification tasks, shedding light on their inherent political biases.

We utilised GPT-3.5, Llama 2-Chat, Mistral, Mistral Chat, and Gemma Instruct for our experiments, employing the same versions as those listed in Table~\ref{tab:generative_models}. Other base models, such as Gemma, Llama 2, and GPT-2, were not used as they were incapable of following the classification steps outlined in the prompt. During the classification of human-written and synthetic news, each model generates a text with a chain-of-thought~\citep{wei2022chain}, culminating in the assigned label within square brackets, which facilitates label extraction. If the classification fails to assign one of the predetermined labels, it gets repeated.

The calculation of the shift follows Equation~\ref{delta}, but without the use of Equation~\ref{val}; instead, we assign the article a value of $-1$ for the left-wing class, $0$ for the centrist, and $1$ for the right-wing.

\subsubsection*{Political Bias Shift Evaluation}
The results of this evaluation, depicted in Figure~\ref{fig:political-shift-LLM}, align with the supervised results in most aspects, confirming the conclusions drawn from our experiment. They demonstrate that instruction-tuned models predominantly exhibit a left-wing bias with a markedly reduced capacity to generate right-wing texts. Furthermore, they validate the general tendency towards left-leaning biases and the neutrality of base models.

In terms of differences, some models exhibit slightly varying strengths of biases, which do not, however, disrupt the general trends observed in supervised models. The most notable changes occur in right-wing scenarios, corresponding to right-wing bias being weaker and less distinct for LLMs, making it more challenging to detect.
\subsubsection*{LLM Classification Behaviour}
\begin{figure}[H]
    \centering
    \includegraphics[width=0.48\textwidth]{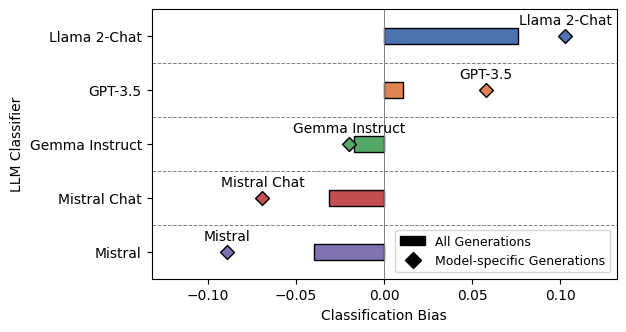}
    \caption{LLM classification bias, with the y-axis denoting the specific model in Equation~\ref{deltaX} and the x-axis the results of the $C_{Bias}(i)$ calculations for two inputs}
    \label{fig:CB}
\end{figure}

As noted, using LLMs as classifiers enables us to investigate the behaviour they exhibit in this role. For this analysis, we employ Equation~\ref{deltaX} to examine classification bias, which measures the difference between the political bias shift calculated by a specific model and the average shift across all models.
\begin{equation}
\label{deltaX}
\mathit{C_{Bias}(i)} = \Delta \mathit{PA}_{\text{specific model}} - \Delta \mathit{PA}_{\text{average}}
\end{equation}
In Figure~\ref{fig:CB}, we compare classification biases across LLMs. For each of them, we calculate two $C_{Bias}(i)$ values using two inputs: one comprising all generated texts, and the other comprising only texts generated by the model conducting the classification. It is apparent from the results that both Gemma Instruct and ChatGPT maintain a relatively centrist position, consistent with previous research~\citep{lin2024investigating}. Models from the Mistral family, notably Mistral Chat, tend to lean left, which aligns with their political bias shift results. The most pronounced behaviour is observed in Llama 2-Chat, which exhibits a distinct bias towards classifying all models, particularly itself, as right-wing.


\section{Discussion}
Through this analysis, we have identified several patterns that offer insights into the behaviour of specific LLMs in generating news content.

\subsection*{Finding 1: Political Alignment of LLMs}
The first observation concerns the political bias evident in some LLMs. Models such as GPT-3.5, GPT-4, Llama 2-Chat, and Mistral Chat demonstrate a notable propensity to produce left-wing content. Other models show a similar, albeit less pronounced, inclination. Our observations corroborate the results of a previous study by~\citet{feng-etal-2023-pretraining}, which reported left-leaning biases in the responses of LLMs to political questionnaires.

\subsection*{Finding 2: Discrepancy Between Instruction-tuned and Base LLMs}
We observe clear evidence of a pronounced discrepancy between articles produced by instruction-tuned and base LLMs. Only the instruction-tuned models exhibited a significant shift in political bias. This could be attributed to the tendency of instruction-tuned models to adopt the writing style of the instruction-tuning data, echoing observations from corresponding studies~\citep{ghosh2024closer}. Thus, this suggests that these models have a more distinct and potentially biased writing style.

\subsection*{Finding 3: Impact of Prompt Types}
The data conclusively shows that prompt types significantly influence bias of generations, with greater effect towards the left. Notably, using the left-wing prompt observably increases the amount of left-leaning articles, whilst in the right-wing context, the shift proves considerably milder, with only a few models demonstrating a rightward tilt and most merely achieving a centrist stance. This trend suggests much lesser resistance within LLMs towards left-leaning stance than for right-wing one.

\subsection*{Finding 4: Classification Behaviour}
When classifying political bias using LLMs, we find that each model attributes to itself a stronger version of the same bias it assigns on average to all models. This suggests that LLMs in a limited way recognise their generations, which influences their assessment. Additionally, while LLMs appear reliable as classifiers, aligning closely with supervised evaluations, certain models, notably Llama 2-Chat and the Mistral family, exhibit pronounced biases. This reveals that political bias in LLMs manifests in multiple facets across their functionalities.

\subsection{Dataset}
To ensure that future examination of this topic is more accessible, the dataset compiled for this study has been made available for any follow-up work.\footnote{\url{https://huggingface.co/datasets/FilipT/Generated_News_Political_Leaning}}

\section{Conclusion}
This study introduces a curated dataset of paired human-written and machine-generated news articles, offering resources and a framework for the quantitative exploration of stylistic and semantic shifts within a journalistic context. We employed this dataset to investigate political bias in LLMs, conducting several experiments with both supervised models and LLM classifiers. Our results reveal significant political bias in instruction-tuned models, raising concerns about their broader application in journalism. Furthermore, the LLMs exhibited a much stronger tendency towards left-wing bias and a greater susceptibility to generating it when prompted, in contrast to right-wing bias. Lastly, we found that LLMs exhibit political bias even in classification tasks, which, alongside findings from other studies, confirms that bias permeates various applications of LLMs. Consequently, we recognise the need for further research before LLMs can be safely integrated into the media industry. We encourage subsequent examinations of LLM behaviour to build on the provided dataset.

\section*{Limitations}
As noted in the study, the detection of political bias remains a challenging problem, with the primary issue being the lack of quality data. Consequently, the risk of models being biased cannot be ignored. This research has a series of measures to mitigate these issues: utilising adjusted values instead of raw data, compiling data from various sources, averaging results across different model architectures, and employing LLMs as classifiers. Nevertheless, although the reported results are well-founded because of these measures, and the reported trend is supported by results from other studies, the exact strengths of biases might vary between this and future research.

Regarding political bias, it is also important to mention that the entire work is heavily centred on the Euro-American political sphere. Both the news dataset and the model training datasets are derived from this context, and the language used for generations was English. Results or entire concepts used in our analysis may differ in other cultural contexts. Moreover, discrepancies may arise within the news categories because the task of condensing the complex subject matter of an article into a single category can lead to erroneous oversimplifications. Nevertheless, this work has provided evidence that a majority of these classifications align with the anticipated outcomes. Consequently, the dataset should be accurate in the vast majority of instances.

In terms of the provided dataset, although we have rigorously tested our prompts, some degenerative patterns remain. We chose not to remove these manually in order to avoid artificially inflating the reported capabilities of the studied LLMs. Critically, these patterns are not intrusive enough to significantly impact our experiments or any future studies using this dataset. Beyond this concern, a further limitation of our dataset is that as advancements in text generation continue, its relevance will likely diminish. Despite this, we expect our dataset to remain useful for several years, which is why we do not view this as significant.

Finally, it must be acknowledged that this dataset was created for the purpose of observing the behaviour of LLMs in a journalistic context. It could also possibly be used to observe other similar properties of generations. However, the generated texts do not aspire to be full-fledged publishable articles, since the length of most generations is too short for that purpose. Therefore, we discourage using this work for any purpose other than its clearly stated main one.

\section*{Ethics Statement}
\paragraph{Interpretation of Political Bias}Defining or quantifying political media bias presents substantial challenges. The results of this study aim to inform the scientific community and contribute to the broader discussion about addressing political bias in LLMs. In our research, specific classes of political bias are derived solely from the training datasets used and the perspectives of the classifying LLMs, without any influence from the authors of this paper \citep{baly-etal-2020-detect, TerAkopyan2021}. Interpretations of left-wing and right-wing leanings can differ significantly among various groups, and we encourage readers to familiarise themselves with the methodologies outlined in the studies associated with these datasets. It is crucial to recognise that the biases identified may not correspond with contemporary political dialogues. Accordingly, this study should not be interpreted as LLMs being aligned or endorsing any given political party.

\paragraph{Misuse Potential}This work identifies which LLMs are prone to generating politically biased content. Our findings are consistent with those already established in the field \citep{feng-etal-2023-pretraining}, and importantly, we do not provide detailed methods that could facilitate misuse. As such, our study should not help any adversarial users in exploiting LLMs. Through this study, we aim to inform the broader community about the biases inherent in these widely accessible models, with the hope of future mitigation of this issue.

\bibliography{custom}

\begin{thebibliography}{63}
\providecommand{\natexlab}[1]{#1}

\bibitem[{Alesina et~al.(2020)Alesina, Miano, and Stantcheva}]{10.1257/pandp.20201072}
Alberto Alesina, Armando Miano, and Stefanie Stantcheva. 2020.
\newblock \href {https://doi.org/10.1257/pandp.20201072} {The polarization of reality}.
\newblock \emph{AEA Papers and Proceedings}, 110:324--28.

\bibitem[{Baly et~al.(2020)Baly, Da~San~Martino, Glass, and Nakov}]{baly-etal-2020-detect}
Ramy Baly, Giovanni Da~San~Martino, James Glass, and Preslav Nakov. 2020.
\newblock \href {https://doi.org/10.18653/v1/2020.emnlp-main.404} {We can detect your bias: Predicting the political ideology of news articles}.
\newblock In \emph{Proceedings of the 2020 Conference on Empirical Methods in Natural Language Processing (EMNLP)}, pages 4982--4991, Online. Association for Computational Linguistics.

\bibitem[{Bird et~al.(2009)Bird, Klein, and Loper}]{bird2009natural}
Steven Bird, Ewan Klein, and Edward Loper. 2009.
\newblock \emph{Natural language processing with Python: analyzing text with the natural language toolkit}.
\newblock " O'Reilly Media, Inc.".

\bibitem[{Chen et~al.(2020)Chen, Al~Khatib, Wachsmuth, and Stein}]{chen-etal-2020-analyzing}
Wei-Fan Chen, Khalid Al~Khatib, Henning Wachsmuth, and Benno Stein. 2020.
\newblock \href {https://doi.org/10.18653/v1/2020.nlpcss-1.16} {Analyzing political bias and unfairness in news articles at different levels of granularity}.
\newblock In \emph{Proceedings of the Fourth Workshop on Natural Language Processing and Computational Social Science}, pages 149--154, Online. Association for Computational Linguistics.

\bibitem[{Christians et~al.(2010)Christians, Glasser, McQuail, Nordenstreng, and White}]{christians2010normative}
Clifford~G Christians, Theodore Glasser, Denis McQuail, Kaarle Nordenstreng, and Robert~A White. 2010.
\newblock \emph{Normative theories of the media: Journalism in democratic societies}.
\newblock University of Illinois Press.

\bibitem[{Dell'Acqua et~al.(2023)Dell'Acqua, McFowland, Mollick, Lifshitz-Assaf, Kellogg, Rajendran, Krayer, Candelon, and Lakhani}]{dellacqua2023navigating}
Fabrizio Dell'Acqua, Edward McFowland, Ethan~R. Mollick, Hila Lifshitz-Assaf, Katherine Kellogg, Saran Rajendran, Lisa Krayer, François Candelon, and Karim~R. Lakhani. 2023.
\newblock Navigating the jagged technological frontier: Field experimental evidence of the effects of ai on knowledge worker productivity and quality.
\newblock Harvard Business School Technology \& Operations Mgt. Unit Working Paper No. 24-013.
\newblock Available at SSRN: https://ssrn.com/abstract=4573321 or http://dx.doi.org/10.2139/ssrn.4573321.

\bibitem[{DellaVigna and Kaplan(2007)}]{10.1162/qjec.122.3.1187}
Stefano DellaVigna and Ethan Kaplan. 2007.
\newblock \href {https://doi.org/10.1162/qjec.122.3.1187} {{The Fox News Effect: Media Bias and Voting*}}.
\newblock \emph{The Quarterly Journal of Economics}, 122(3):1187--1234.

\bibitem[{Elejalde et~al.(2018)Elejalde, Ferres, and Herder}]{RePEc:plo:pone00:0193765}
Erick Elejalde, Leo Ferres, and Eelco Herder. 2018.
\newblock \href {https://doi.org/10.1371/journal.pone.0193} {{On the nature of real and perceived bias in the mainstream media}}.
\newblock \emph{PLOS ONE}, 13(3):1--28.

\bibitem[{Fecher et~al.(2023)Fecher, Hebing, Laufer, Pohle, and Sofsky}]{Fecher2023FriendOF}
Benedikt Fecher, Marcel Hebing, Melissa Laufer, J{\"o}rg Pohle, and Fabian Sofsky. 2023.
\newblock \href {https://doi.org/10.1007/s00146-023-01791-1} {Friend or foe? exploring the implications of large language models on the science system}.
\newblock \emph{AI \& SOCIETY}, 10.1007/s00146-023-01791-1.

\bibitem[{Feng et~al.(2021)Feng, Chen, Zhang, Li, Zheng, Chang, and Luo}]{feng2021kgap}
Shangbin Feng, Zilong Chen, Wenqian Zhang, Qingyao Li, Qinghua Zheng, Xiaojun Chang, and Minnan Luo. 2021.
\newblock Kgap: Knowledge graph augmented political perspective detection in news media.
\newblock \emph{arXiv preprint arXiv:2108.03861}.

\bibitem[{Feng et~al.(2023)Feng, Park, Liu, and Tsvetkov}]{feng-etal-2023-pretraining}
Shangbin Feng, Chan~Young Park, Yuhan Liu, and Yulia Tsvetkov. 2023.
\newblock \href {https://doi.org/10.18653/v1/2023.acl-long.656} {From pretraining data to language models to downstream tasks: Tracking the trails of political biases leading to unfair {NLP} models}.
\newblock In \emph{Proceedings of the 61st Annual Meeting of the Association for Computational Linguistics (Volume 1: Long Papers)}, pages 11737--11762, Toronto, Canada. Association for Computational Linguistics.

\bibitem[{Gallegos et~al.(2023)Gallegos, Rossi, Barrow, Tanjim, Kim, Dernoncourt, Yu, Zhang, and Ahmed}]{Gallegos2023BiasAF}
Isabel~O. Gallegos, Ryan~A. Rossi, Joe Barrow, Md.~Mehrab Tanjim, Sungchul Kim, Franck Dernoncourt, Tong Yu, Ruiyi Zhang, and Nesreen Ahmed. 2023.
\newblock \href {https://api.semanticscholar.org/CorpusID:261530629} {Bias and fairness in large language models: A survey}.
\newblock \emph{ArXiv}, abs/2309.00770.

\bibitem[{Ganguli et~al.(2023)Ganguli, Askell, Schiefer, Liao, Luko{\v{s}}i{\=u}t{\.e}, Chen, Goldie, Mirhoseini, Olsson, Hernandez et~al.}]{ganguli2023capacity}
Deep Ganguli, Amanda Askell, Nicholas Schiefer, Thomas~I Liao, Kamil{\.e} Luko{\v{s}}i{\=u}t{\.e}, Anna Chen, Anna Goldie, Azalia Mirhoseini, Catherine Olsson, Danny Hernandez, et~al. 2023.
\newblock The capacity for moral self-correction in large language models.
\newblock \emph{arXiv preprint arXiv:2302.07459}.

\bibitem[{Gemma~Team et~al.(2024)Gemma~Team, Mesnard, Hardin, Dadashi, Bhupatiraju, Pathak, Sifre, Rivi{\`e}re, Kale, Love et~al.}]{gemma_2024}
Gemma Gemma~Team, Thomas Mesnard, Cassidy Hardin, Robert Dadashi, Surya Bhupatiraju, Shreya Pathak, Laurent Sifre, Morgane Rivi{\`e}re, Mihir~Sanjay Kale, Juliette Love, et~al. 2024.
\newblock Gemma: Open models based on gemini research and technology.
\newblock \emph{arXiv preprint arXiv:2403.08295}.

\bibitem[{Ghosh et~al.(2024)Ghosh, Evuru, Kumar, Aneja, Jin, Duraiswami, Manocha et~al.}]{ghosh2024closer}
Sreyan Ghosh, Chandra Kiran~Reddy Evuru, Sonal Kumar, Deepali Aneja, Zeyu Jin, Ramani Duraiswami, Dinesh Manocha, et~al. 2024.
\newblock A closer look at the limitations of instruction tuning.
\newblock \emph{arXiv preprint arXiv:2402.05119}.

\bibitem[{Gold(1998)}]{Gold_1998}
Thomas~W. Gold. 1998.
\newblock \href {https://doi.org/10.2307/2585939} {Left and right: The significance of a political distinction. by noberto bobbio. chicago: University of chicago press. 1997. 148p. \$34.95 cloth, \$14.95 paper.}
\newblock \emph{American Political Science Review}, 92(1):199–199.

\bibitem[{Grusky et~al.(2018)Grusky, Naaman, and Artzi}]{grusky-etal-2018-newsroom}
Max Grusky, Mor Naaman, and Yoav Artzi. 2018.
\newblock \href {https://doi.org/10.18653/v1/N18-1065} {{N}ewsroom: A dataset of 1.3 million summaries with diverse extractive strategies}.
\newblock In \emph{Proceedings of the 2018 Conference of the North {A}merican Chapter of the Association for Computational Linguistics: Human Language Technologies, Volume 1 (Long Papers)}, pages 708--719, New Orleans, Louisiana. Association for Computational Linguistics.

\bibitem[{Guess and Lyons(2020)}]{Guess_Lyons_2020}
Andrew~M. Guess and Benjamin~A. Lyons. 2020.
\newblock \emph{Misinformation, Disinformation, and Online Propaganda}, page 10–33.
\newblock SSRC Anxieties of Democracy. Cambridge University Press.

\bibitem[{Harris et~al.(2020)Harris, Millman, Van Der~Walt, Gommers, Virtanen, Cournapeau, Wieser, Taylor, Berg, Smith et~al.}]{harris2020array}
Charles~R Harris, K~Jarrod Millman, St{\'e}fan~J Van Der~Walt, Ralf Gommers, Pauli Virtanen, David Cournapeau, Eric Wieser, Julian Taylor, Sebastian Berg, Nathaniel~J Smith, et~al. 2020.
\newblock Array programming with numpy.
\newblock \emph{Nature}, 585(7825):357--362.

\bibitem[{Holtzman et~al.(2019)Holtzman, Buys, Du, Forbes, and Choi}]{holtzman2019curious}
Ari Holtzman, Jan Buys, Li~Du, Maxwell Forbes, and Yejin Choi. 2019.
\newblock The curious case of neural text degeneration.
\newblock \emph{arXiv preprint arXiv:1904.09751}.

\bibitem[{Humlum and Vestergaard(2024)}]{humlum2024adoption}
Anders Humlum and Emilie Vestergaard. 2024.
\newblock \href {https://doi.org/10.2139/ssrn.4807516} {The adoption of chatgpt}.
\newblock Technical Report 2024-50, University of Chicago, Becker Friedman Institute for Economics.

\bibitem[{Iyyer et~al.(2014)Iyyer, Enns, Boyd-Graber, and Resnik}]{iyyer-etal-2014-political}
Mohit Iyyer, Peter Enns, Jordan Boyd-Graber, and Philip Resnik. 2014.
\newblock \href {https://doi.org/10.3115/v1/P14-1105} {Political ideology detection using recursive neural networks}.
\newblock In \emph{Proceedings of the 52nd Annual Meeting of the Association for Computational Linguistics (Volume 1: Long Papers)}, pages 1113--1122, Baltimore, Maryland. Association for Computational Linguistics.

\bibitem[{Jashinsky et~al.(2016)Jashinsky, Magnusson, Hanson, and Barnes}]{Jashinsky2016}
Jared~Michael Jashinsky, Brianna Magnusson, Carl Hanson, and Michael Barnes. 2016.
\newblock \href {https://doi.org/10.3389/fpubh.2016.00291} {Media agenda setting regarding gun violence before and after a mass shooting}.
\newblock \emph{Frontiers in Public Health}, 4:291.

\bibitem[{Jiang et~al.(2023)Jiang, Sablayrolles, Mensch, Bamford, Chaplot, de~las Casas, Bressand, Lengyel, Lample, Saulnier, Lavaud, Lachaux, Stock, Scao, Lavril, Wang, Lacroix, and Sayed}]{jiang2023mistral}
Albert~Q. Jiang, Alexandre Sablayrolles, Arthur Mensch, Chris Bamford, Devendra~Singh Chaplot, Diego de~las Casas, Florian Bressand, Gianna Lengyel, Guillaume Lample, Lucile Saulnier, Lélio~Renard Lavaud, Marie-Anne Lachaux, Pierre Stock, Teven~Le Scao, Thibaut Lavril, Thomas Wang, Timothée Lacroix, and William~El Sayed. 2023.
\newblock \href {https://arxiv.org/abs/2310.06825} {Mistral 7b}.
\newblock \emph{Preprint}, arXiv:2310.06825.

\bibitem[{Keskar et~al.(2019)Keskar, McCann, Varshney, Xiong, and Socher}]{keskar2019ctrl}
Nitish~Shirish Keskar, Bryan McCann, Lav~R. Varshney, Caiming Xiong, and Richard Socher. 2019.
\newblock \href {https://arxiv.org/abs/1909.05858} {Ctrl: A conditional transformer language model for controllable generation}.
\newblock \emph{Preprint}, arXiv:1909.05858.

\bibitem[{Kulkarni et~al.(2018)Kulkarni, Ye, Skiena, and Wang}]{kulkarni-etal-2018-multi}
Vivek Kulkarni, Junting Ye, Steve Skiena, and William~Yang Wang. 2018.
\newblock \href {https://doi.org/10.18653/v1/D18-1388} {Multi-view models for political ideology detection of news articles}.
\newblock In \emph{Proceedings of the 2018 Conference on Empirical Methods in Natural Language Processing}, pages 3518--3527, Brussels, Belgium. Association for Computational Linguistics.

\bibitem[{Lane(1956)}]{Lane_1956}
Robert~E. Lane. 1956.
\newblock \href {https://doi.org/10.1017/S000305540023027X} {The psychology of politics. by h. j. eysenck. (new york: Frederick a. praeger. 1955. pp. xvi, 317. \$6.00.)}.
\newblock \emph{American Political Science Review}, 50(2):599–599.

\bibitem[{Lang(1995)}]{Lang95}
Ken Lang. 1995.
\newblock Newsweeder: Learning to filter netnews.
\newblock In \emph{Proceedings of the Twelfth International Conference on Machine Learning}, pages 331--339.

\bibitem[{Legagneux et~al.(2018)Legagneux, Casajus, Cazelles, Chevallier, Chevrinais, Guéry, Jacquet, Jaffré, Naud, Noisette, Ropars, Vissault, Archambault, Bêty, Berteaux, and Gravel}]{10.3389/fevo.2017.00175}
Pierre Legagneux, Nicolas Casajus, Kevin Cazelles, Clément Chevallier, Marion Chevrinais, Lorelei Guéry, Claire Jacquet, Mikael Jaffré, Marie-José Naud, Fanny Noisette, Pascale Ropars, Steve Vissault, Philippe Archambault, Joel Bêty, Dominique Berteaux, and Dominique Gravel. 2018.
\newblock \href {https://doi.org/10.3389/fevo.2017.00175} {Our house is burning: Discrepancy in climate change vs. biodiversity coverage in the media as compared to scientific literature}.
\newblock \emph{Frontiers in Ecology and Evolution}, 5.

\bibitem[{Lin et~al.(2024{\natexlab{a}})Lin, Wang, Guo, and Wong}]{lin2024investigating}
Luyang Lin, Lingzhi Wang, Jinsong Guo, and Kam-Fai Wong. 2024{\natexlab{a}}.
\newblock Investigating bias in llm-based bias detection: Disparities between llms and human perception.
\newblock \emph{arXiv preprint arXiv:2403.14896}.

\bibitem[{Lin et~al.(2024{\natexlab{b}})Lin, Wang, Zhao, Li, and Wong}]{lin-etal-2024-indivec}
Luyang Lin, Lingzhi Wang, Xiaoyan Zhao, Jing Li, and Kam-Fai Wong. 2024{\natexlab{b}}.
\newblock \href {https://aclanthology.org/2024.findings-eacl.70} {{I}ndi{V}ec: An exploration of leveraging large language models for media bias detection with fine-grained bias indicators}.
\newblock In \emph{Findings of the Association for Computational Linguistics: EACL 2024}, pages 1038--1050, St. Julian{'}s, Malta. Association for Computational Linguistics.

\bibitem[{Liu et~al.(2019)Liu, Ott, Goyal, Du, Joshi, Chen, Levy, Lewis, Zettlemoyer, and Stoyanov}]{liu2019roberta}
Yinhan Liu, Myle Ott, Naman Goyal, Jingfei Du, Mandar Joshi, Danqi Chen, Omer Levy, Mike Lewis, Luke Zettlemoyer, and Veselin Stoyanov. 2019.
\newblock Roberta: A robustly optimized bert pretraining approach.
\newblock \emph{arXiv preprint arXiv:1907.11692}.

\bibitem[{Liu et~al.(2022)Liu, Zhang, Wegsman, Beauchamp, and Wang}]{liu-etal-2022-politics}
Yujian Liu, Xinliang~Frederick Zhang, David Wegsman, Nicholas Beauchamp, and Lu~Wang. 2022.
\newblock \href {https://doi.org/10.18653/v1/2022.findings-naacl.101} {{POLITICS}: Pretraining with same-story article comparison for ideology prediction and stance detection}.
\newblock In \emph{Findings of the Association for Computational Linguistics: NAACL 2022}, pages 1354--1374, Seattle, United States. Association for Computational Linguistics.

\bibitem[{Martens et~al.(2018)Martens, Aguiar, Gomez-Herrera, and Mueller-Langer}]{martens2018digital}
Bertin Martens, Luis Aguiar, Estrella Gomez-Herrera, and Frank Mueller-Langer. 2018.
\newblock \href {https://doi.org/10.2139/ssrn.3164170} {The digital transformation of news media and the rise of disinformation and fake news}.
\newblock Digital Economy Working Paper 2018-02, Joint Research Centre.

\bibitem[{Mckinney(2011)}]{pandas}
Wes Mckinney. 2011.
\newblock pandas: a foundational python library for data analysis and statistics.
\newblock \emph{Python High Performance Science Computer}.

\bibitem[{Misra(2022)}]{misra2022news}
Rishabh Misra. 2022.
\newblock News category dataset.
\newblock \emph{arXiv preprint arXiv:2209.11429}.

\bibitem[{Morgan(2013)}]{doi:10.1177/2050303213506476}
David Morgan. 2013.
\newblock \href {https://doi.org/10.1177/2050303213506476} {Religion and media: A critical review of recent developments}.
\newblock \emph{Critical Research on Religion}, 1(3):347--356.

\bibitem[{Motoki et~al.(2023)Motoki, Pinho~Neto, and Rangel}]{Motoki2023}
Fabio Motoki, Valdemar Pinho~Neto, and Victor Rangel. 2023.
\newblock \href {https://doi.org/10.1007/s11127-023-01097-2} {More human than human: measuring chatgpt political bias}.
\newblock \emph{Public Choice}, 198.

\bibitem[{Newman et~al.(2023)Newman, Fletcher, Eddy, Robertson, and Nielsen}]{newman2023a}
N~Newman, R~Fletcher, K~Eddy, CT~Robertson, and RK~Nielsen. 2023.
\newblock \href {https://reutersinstitute.politics.ox.ac.uk/sites/default/files/2023-01/Journalism_media_and_technology_trends_and_predictions_2023.pdf} {Digital news report 2023}.
\newblock Technical report.

\bibitem[{Nishal and Diakopoulos(2024)}]{nishal2024envisioning}
Sachita Nishal and Nicholas Diakopoulos. 2024.
\newblock Envisioning the applications and implications of generative ai for news media.
\newblock \emph{arXiv preprint arXiv:2402.18835}.

\bibitem[{Nolan(2019)}]{Nolan2019}
Scott~N. Nolan. 2019.
\newblock \href {https://scholarworks.uno.edu/td/2629} {Media coverage of lgbt issues: Legal, religious, and political frames}.
\newblock Theses and dissertations, University of New Orleans.

\bibitem[{Ntoutsi et~al.(2020)Ntoutsi, Fafalios, Gadiraju, Iosifidis, Nejdl, Vidal, Ruggieri, Turini, Papadopoulos, Krasanakis, Kompatsiaris, Kinder-Kurlanda, Wagner, Karimi, Fernandez, Alani, Berendt, Kruegel, Heinze, Broelemann, Kasneci, Tiropanis, and Staab}]{ntoutsi2020}
Eirini Ntoutsi, Pavlos Fafalios, Ujwal Gadiraju, Vasileios Iosifidis, Wolfgang Nejdl, Maria-Esther Vidal, Salvatore Ruggieri, Franco Turini, Symeon Papadopoulos, Emmanouil Krasanakis, Ioannis Kompatsiaris, Katharina Kinder-Kurlanda, Claudia Wagner, Fariba Karimi, Miriam Fernandez, Harith Alani, Bettina Berendt, Tina Kruegel, Christian Heinze, Klaus Broelemann, Gjergji Kasneci, Thanassis Tiropanis, and Steffen Staab. 2020.
\newblock \href {https://doi.org/10.1002/widm.1356} {Bias in data-driven artificial intelligence systems—an introductory survey}.
\newblock \emph{WIREs Data Mining and Knowledge Discovery}, 10(3):e1356.

\bibitem[{OpenAI(2022)}]{chatgpt-etal-2022}
OpenAI. 2022.
\newblock \href {https://openai.com/blog/chatgpt} {{ChatGPT: Optimizing Language Models for Dialogue}}.

\bibitem[{OpenAI(2023)}]{openai2023gpt4}
OpenAI. 2023.
\newblock \href {https://arxiv.org/abs/2303.08774} {Gpt-4 technical report}.
\newblock \emph{Preprint}, arXiv:2303.08774.

\bibitem[{Ouyang et~al.(2022)Ouyang, Wu, Jiang, Almeida, Wainwright, Mishkin, Zhang, Agarwal, Slama, Ray, Schulman, Hilton, Kelton, Miller, Simens, Askell, Welinder, Christiano, Leike, and Lowe}]{Ouyang2022TrainingLM}
Long Ouyang, Jeff Wu, Xu~Jiang, Diogo Almeida, Carroll~L. Wainwright, Pamela Mishkin, Chong Zhang, Sandhini Agarwal, Katarina Slama, Alex Ray, John Schulman, Jacob Hilton, Fraser Kelton, Luke~E. Miller, Maddie Simens, Amanda Askell, Peter Welinder, Paul~Francis Christiano, Jan Leike, and Ryan~J. Lowe. 2022.
\newblock \href {https://api.semanticscholar.org/CorpusID:246426909} {Training language models to follow instructions with human feedback}.
\newblock \emph{ArXiv}, abs/2203.02155.

\bibitem[{Paszke et~al.(2019)Paszke, Gross, Massa, Lerer, Bradbury, Chanan, Killeen, Lin, Gimelshein, Antiga et~al.}]{paszke2019pytorch}
Adam Paszke, Sam Gross, Francisco Massa, Adam Lerer, James Bradbury, Gregory Chanan, Trevor Killeen, Zeming Lin, Natalia Gimelshein, Luca Antiga, et~al. 2019.
\newblock Pytorch: An imperative style, high-performance deep learning library.
\newblock \emph{Advances in neural information processing systems}, 32.

\bibitem[{Radford et~al.(2019)Radford, Wu, Child, Luan, Amodei, and Sutskever}]{Radford2019LanguageMA}
Alec Radford, Jeff Wu, Rewon Child, David Luan, Dario Amodei, and Ilya Sutskever. 2019.
\newblock \href {https://api.semanticscholar.org/CorpusID:160025533} {Language models are unsupervised multitask learners}.

\bibitem[{Russo(2023)}]{Russo2023NavigatingTC}
Daniel Russo. 2023.
\newblock \href {https://api.semanticscholar.org/CorpusID:259836763} {Navigating the complexity of generative ai adoption in software engineering}.
\newblock \emph{ArXiv}, abs/2307.06081.

\bibitem[{Sarvazyan et~al.(2023)Sarvazyan, González, Rosso, and Franco~Salvador}]{Sarvazyan2023}
Areg Sarvazyan, José González, Paolo Rosso, and Marc Franco~Salvador. 2023.
\newblock \href {https://doi.org/10.1007/978-3-031-42448-9_11} {\emph{Supervised Machine-Generated Text Detectors: Family and Scale Matters}}, pages 121--132.

\bibitem[{Sennrich et~al.(2016)Sennrich, Haddow, and Birch}]{sennrich-etal-2016-neural}
Rico Sennrich, Barry Haddow, and Alexandra Birch. 2016.
\newblock \href {https://doi.org/10.18653/v1/P16-1162} {Neural machine translation of rare words with subword units}.
\newblock In \emph{Proceedings of the 54th Annual Meeting of the Association for Computational Linguistics (Volume 1: Long Papers)}, pages 1715--1725, Berlin, Germany. Association for Computational Linguistics.

\bibitem[{Shanahan and Clarke(2023)}]{Shanahan2023EvaluatingLL}
Murray Shanahan and Catherine Clarke. 2023.
\newblock \href {https://api.semanticscholar.org/CorpusID:266053546} {Evaluating large language model creativity from a literary perspective}.
\newblock \emph{ArXiv}, abs/2312.03746.

\bibitem[{Spinde et~al.(2021)Spinde, Plank, Krieger, Ruas, Gipp, and Aizawa}]{spinde-etal-2021-neural-media}
Timo Spinde, Manuel Plank, Jan-David Krieger, Terry Ruas, Bela Gipp, and Akiko Aizawa. 2021.
\newblock \href {https://doi.org/10.18653/v1/2021.findings-emnlp.101} {Neural media bias detection using distant supervision with {BABE} - bias annotations by experts}.
\newblock In \emph{Findings of the Association for Computational Linguistics: EMNLP 2021}, pages 1166--1177, Punta Cana, Dominican Republic. Association for Computational Linguistics.

\bibitem[{Sun et~al.(2023)Sun, Li, Li, Wu, Guo, Zhang, and Wang}]{sun2023text}
Xiaofei Sun, Xiaoya Li, Jiwei Li, Fei Wu, Shangwei Guo, Tianwei Zhang, and Guoyin Wang. 2023.
\newblock Text classification via large language models.
\newblock \emph{arXiv preprint arXiv:2305.08377}.

\bibitem[{Ter-Akopyan(2021)}]{TerAkopyan2021}
Bagrat Ter-Akopyan. 2021.
\newblock \href {https://www.en.pms.ifi.lmu.de/publications/diplomarbeiten/Bagrat.Ter-Akopyan/MA_Bagrat.Ter-Akopyan.pdf} {Identification of political leaning in german news articles}.
\newblock Master's thesis, Ludwig Maximilian University of Munich, Munich, Germany, 7.
\newblock Retrieved from Ludwig Maximilian University of Munich, Department of Informatics.

\bibitem[{Touvron et~al.(2023)Touvron, Martin, Stone, Albert, Almahairi, Babaei, Bashlykov, Batra, Bhargava, Bhosale, Bikel, Blecher, Ferrer, Chen, Cucurull, Esiobu, Fernandes, Fu, Fu, Fuller, Gao, Goswami, Goyal, Hartshorn, Hosseini, Hou, Inan, Kardas, Kerkez, Khabsa, Kloumann, Korenev, Koura, Lachaux, Lavril, Lee, Liskovich, Lu, Mao, Martinet, Mihaylov, Mishra, Molybog, Nie, Poulton, Reizenstein, Rungta, Saladi, Schelten, Silva, Smith, Subramanian, Tan, Tang, Taylor, Williams, Kuan, Xu, Yan, Zarov, Zhang, Fan, Kambadur, Narang, Rodriguez, Stojnic, Edunov, and Scialom}]{touvron2023llama}
Hugo Touvron, Louis Martin, Kevin Stone, Peter Albert, Amjad Almahairi, Yasmine Babaei, Nikolay Bashlykov, Soumya Batra, Prajjwal Bhargava, Shruti Bhosale, Dan Bikel, Lukas Blecher, Cristian~Canton Ferrer, Moya Chen, Guillem Cucurull, David Esiobu, Jude Fernandes, Jeremy Fu, Wenyin Fu, Brian Fuller, Cynthia Gao, Vedanuj Goswami, Naman Goyal, Anthony Hartshorn, Saghar Hosseini, Rui Hou, Hakan Inan, Marcin Kardas, Viktor Kerkez, Madian Khabsa, Isabel Kloumann, Artem Korenev, Punit~Singh Koura, Marie-Anne Lachaux, Thibaut Lavril, Jenya Lee, Diana Liskovich, Yinghai Lu, Yuning Mao, Xavier Martinet, Todor Mihaylov, Pushkar Mishra, Igor Molybog, Yixin Nie, Andrew Poulton, Jeremy Reizenstein, Rashi Rungta, Kalyan Saladi, Alan Schelten, Ruan Silva, Eric~Michael Smith, Ranjan Subramanian, Xiaoqing~Ellen Tan, Binh Tang, Ross Taylor, Adina Williams, Jian~Xiang Kuan, Puxin Xu, Zheng Yan, Iliyan Zarov, Yuchen Zhang, Angela Fan, Melanie Kambadur, Sharan Narang, Aurelien Rodriguez, Robert Stojnic, Sergey Edunov, and Thomas
  Scialom. 2023.
\newblock \href {https://arxiv.org/abs/2307.09288} {Llama 2: Open foundation and fine-tuned chat models}.
\newblock \emph{Preprint}, arXiv:2307.09288.

\bibitem[{Unanue et~al.(2023)Unanue, Haffari, and Piccardi}]{10.1162/tacl_a_00593}
Inigo~Jauregi Unanue, Gholamreza Haffari, and Massimo Piccardi. 2023.
\newblock \href {https://doi.org/10.1162/tacl_a_00593} {{T3L: Translate-and-Test Transfer Learning for Cross-Lingual Text Classification}}.
\newblock \emph{Transactions of the Association for Computational Linguistics}, 11:1147--1161.

\bibitem[{Urman and Makhortykh(2023)}]{urman_makhortykh_2023}
Aleksandra Urman and Mykola Makhortykh. 2023.
\newblock \href {https://doi.org/10.31219/osf.io/q9v8f} {The silence of the llms: Cross-lingual analysis of political bias and false information prevalence in chatgpt, google bard, and bing chat}.

\bibitem[{{van Giffen} et~al.(2022){van Giffen}, Herhausen, and Fahse}]{VANGIFFEN202293}
Benjamin {van Giffen}, Dennis Herhausen, and Tobias Fahse. 2022.
\newblock \href {https://doi.org/10.1016/j.jbusres.2022.01.076} {Overcoming the pitfalls and perils of algorithms: A classification of machine learning biases and mitigation methods}.
\newblock \emph{Journal of Business Research}, 144:93--106.

\bibitem[{Wei et~al.(2022)Wei, Wang, Schuurmans, Bosma, Xia, Chi, Le, Zhou et~al.}]{wei2022chain}
Jason Wei, Xuezhi Wang, Dale Schuurmans, Maarten Bosma, Fei Xia, Ed~Chi, Quoc~V Le, Denny Zhou, et~al. 2022.
\newblock Chain-of-thought prompting elicits reasoning in large language models.
\newblock \emph{Advances in neural information processing systems}, 35:24824--24837.

\bibitem[{Wiher et~al.(2022)Wiher, Meister, and Cotterell}]{wiher2022decoding}
Gian Wiher, Clara Meister, and Ryan Cotterell. 2022.
\newblock On decoding strategies for neural text generators.
\newblock \emph{Transactions of the Association for Computational Linguistics}, 10:997--1012.

\bibitem[{Wolf et~al.(2020)Wolf, Debut, Sanh, Chaumond, Delangue, Moi, Cistac, Rault, Louf, Funtowicz, Davison, Shleifer, von Platen, Ma, Jernite, Plu, Xu, Le~Scao, Gugger, Drame, Lhoest, and Rush}]{wolf-etal-2020-transformers}
Thomas Wolf, Lysandre Debut, Victor Sanh, Julien Chaumond, Clement Delangue, Anthony Moi, Pierric Cistac, Tim Rault, Remi Louf, Morgan Funtowicz, Joe Davison, Sam Shleifer, Patrick von Platen, Clara Ma, Yacine Jernite, Julien Plu, Canwen Xu, Teven Le~Scao, Sylvain Gugger, Mariama Drame, Quentin Lhoest, and Alexander Rush. 2020.
\newblock \href {https://doi.org/10.18653/v1/2020.emnlp-demos.6} {Transformers: State-of-the-art natural language processing}.
\newblock In \emph{Proceedings of the 2020 Conference on Empirical Methods in Natural Language Processing: System Demonstrations}, pages 38--45, Online. Association for Computational Linguistics.

\bibitem[{Zhang et~al.(2022)Zhang, Feng, Chen, Lei, Li, and Luo}]{zhang-etal-2022-kcd}
Wenqian Zhang, Shangbin Feng, Zilong Chen, Zhenyu Lei, Jundong Li, and Minnan Luo. 2022.
\newblock \href {https://doi.org/10.18653/v1/2022.naacl-main.304} {{KCD}: Knowledge walks and textual cues enhanced political perspective detection in news media}.
\newblock In \emph{Proceedings of the 2022 Conference of the North American Chapter of the Association for Computational Linguistics: Human Language Technologies}, pages 4129--4140, Seattle, United States. Association for Computational Linguistics.

\bibitem[{Zhu et~al.(2018)Zhu, Lu, Zheng, Guo, Zhang, Wang, and Yu}]{zhu2018texygen}
Yaoming Zhu, Sidi Lu, Lei Zheng, Jiaxian Guo, Weinan Zhang, Jun Wang, and Yong Yu. 2018.
\newblock Texygen: A benchmarking platform for text generation models.
\newblock In \emph{The 41st international ACM SIGIR conference on research \& development in information retrieval}, pages 1097--1100.

\end{thebibliography}

\appendix

\section{Hyperparameter Settings \& Specifications}
\begin{table}[H]
\centering
\begin{tabular}{p{1.5cm}|p{2.5cm}|p{2cm}}
\hline
\textbf{Category}   & \textbf{20 Newsgroup} & \textbf{News Category Dataset} \\ \hline
Politics & Politics        & Politics              \\
Guns     & Guns            & -                   \\
LGBT     & -             & Queer voices         \\
Religion & Religion        & Religion              \\
Business & -             & Business              \\
Ecology  & -             & Green                 \\
Sports   & Sport           & Sports                \\ \hline
\end{tabular}
\caption{Category assignment based on 20 Newsgroup and News Category Dataset labels}
\label{tab:category_assignment}
\end{table}

\begin{table}[H]
\centering
\begin{tabular}{l|l}
\hline
\textbf{Hyperparameter} & \textbf{Value} \\
\hline
number of epochs & 10 \\
patience & 4 \\
learning rate & \(1e^{-5}\) \\
optimizer & AdamW \\
max input size & 512 \\
\hline
\end{tabular}
\caption{Model hyperparameters}
\label{table:hyperparameters}
\end{table}

\begin{table}[h]
\centering
\begin{tabular}{l|l}
\hline
\textbf{Hyperparameter} & \textbf{Value} \\
\hline
Temperature & 1 \\
Top-p & 1 \\
Generation length & 512 \\
Repetition penalty & 1 or 1.2 \\
\hline
\end{tabular}
\caption{Generation hyperparameters}
\label{table:hyperparameters_gen}
\end{table}

\begin{table}[H]
\centering
    \begin{tabular}{p{3.75cm}|p{3.2cm}} 
    \hline
         \textbf{Instruction} & \textbf{Continuous} \\ 
         \hline
         \small{This is a political bias classifier for news articles. First, list CLUES (such as keywords, phrases, contextual information, semantic relations, semantic meanings, tones, and references) that support the determination of political bias in the input. Second, deduce the diagnostic REASONING process from the premises (i.e., clues and input) that supports the determination of political bias in the INPUT (limit the number of words to 130). Third, based on the clues, reasoning, and input, determine the overall political bias of the INPUT as "right-wing", "centrist", or "left-wing". Write the FINAL\_LABEL as a single word in brackets [] at the end of the text. INPUT: "\{text\}"}
        & \small{INPUT: "\{text\}" This is the political bias classification of the given news article. First, it lists CLUES that support the determination of political bias in the INPUT. Second, it deduces the diagnostic REASONING process from the premises (i.e., clues, input) that supports the determination of political bias in the INPUT. Third, based on the clues, reasoning, and input, it determines the overall political bias and writes the FINAL\_LABEL as a single word in brackets [], it can be "right-wing," "centrist," or "left-wing." Here are the CLUES, REASONING, and FINAL\_LABEL:}\\ 
         \hline
    \end{tabular}
    \caption{Classification prompts}
    \label{tab:CARP}
\end{table}
\clearpage

\section{Dataset Overview}
\begin{figure*}[h]
    \centering
    \includegraphics[width=1.0\textwidth]{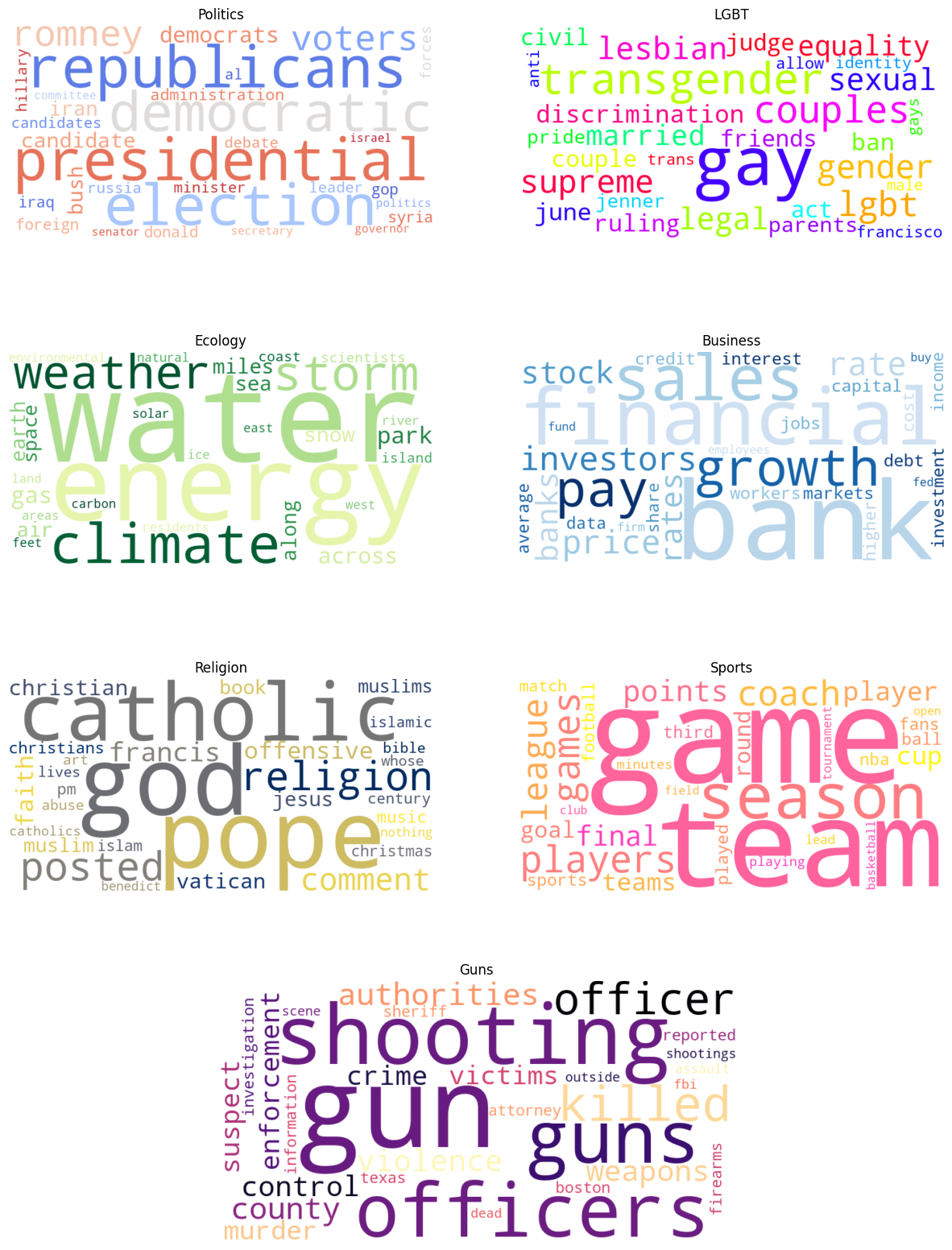}
    \caption{Most frequent unique words per category}
    \label{fig:WC_groups}
\end{figure*}

\begin{figure*}[h]
    \centering
    \includegraphics[width=1.0\textwidth]{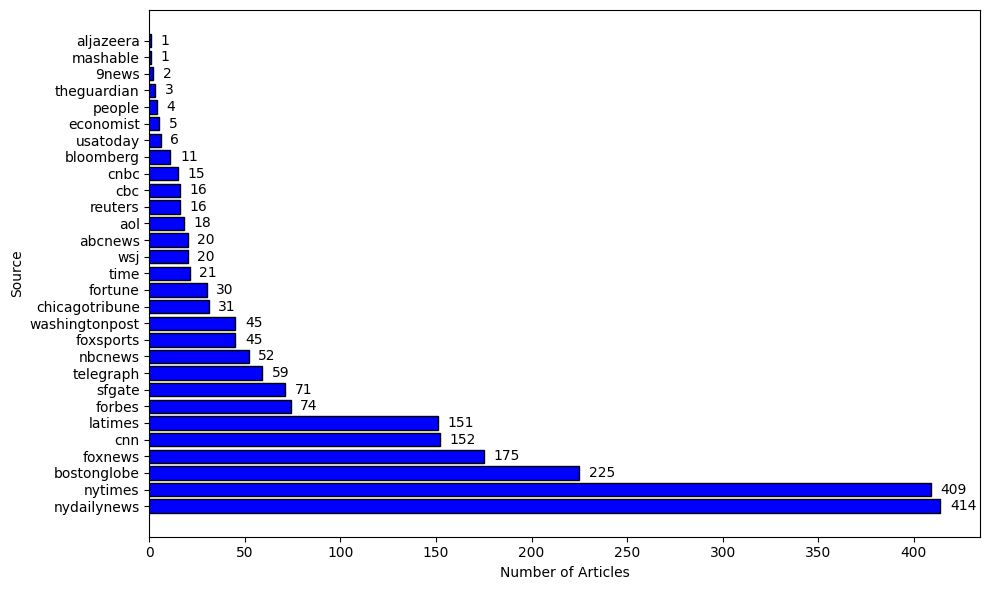}
    \caption{News source distribution in the final dataset}
    \label{fig:num_of_sources}
\end{figure*}

\begin{figure*}[h]
    \centering
    \includegraphics[width=1.0\textwidth]{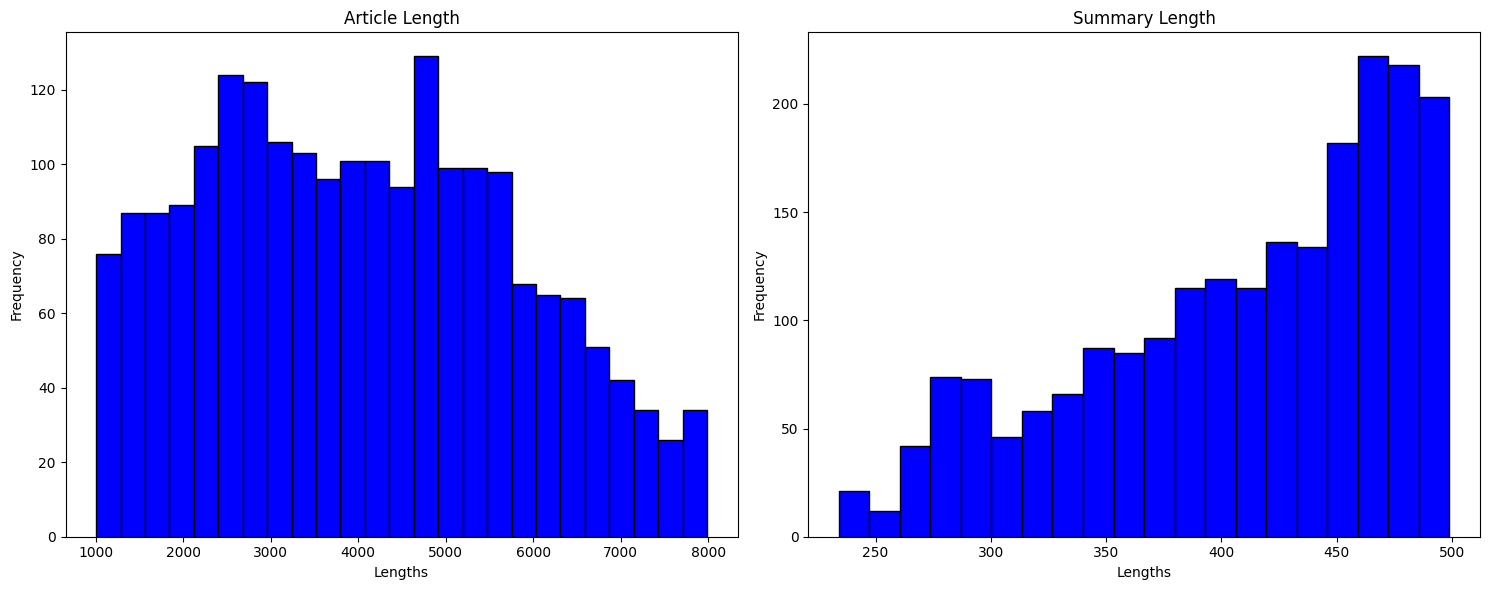}
    \caption{Frequency of articles and summaries in the final dataset based on their length}
    \label{fig:2hist}
\end{figure*}

\begin{figure*}[h]
    \centering
    \includegraphics[width=1.0\textwidth]{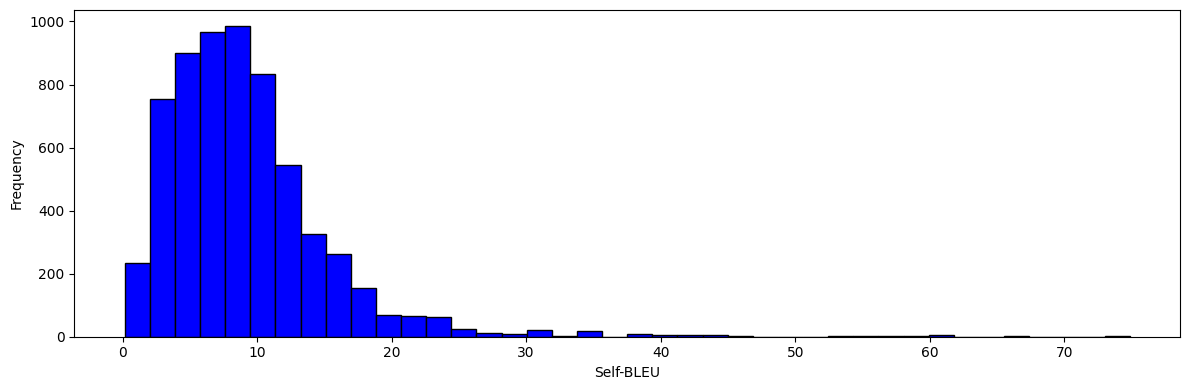}
    \caption{Frequency of human-written articles per Self-BLEU value}
    \label{fig:humanbleu}
\end{figure*}

\begin{figure*}[h]
    \centering
    \includegraphics[width=1.0\textwidth]{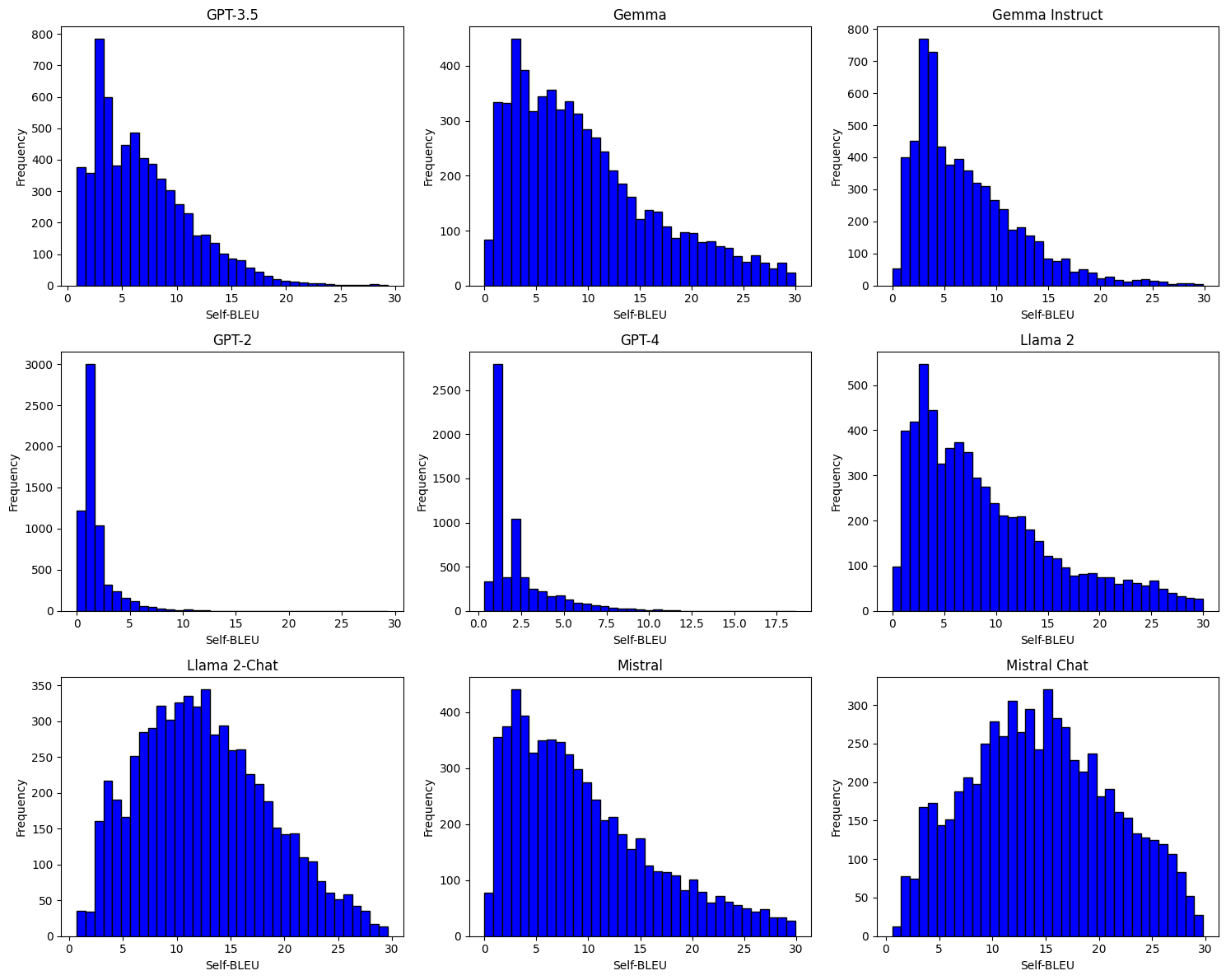}
    \caption{Frequency of articles per Self-BLEU value with respect to specific models}
    \label{fig:bleu}
\end{figure*}

The finished dataset contains 56,700 rows and 46 columns. Here, we provide descriptions of all the columns to assist readers in understanding its structure.
\paragraph{text} Text of the human-written article
\paragraph{model} LLM used for generation
\paragraph{prompt} Prompt used for generation
\paragraph{title} Title of the human-written news article
\paragraph{news\_category} News category label of the human-written news article
\paragraph{main\_domain} Source web domain from which the article originates
\paragraph{length\_human} Length of the human-authored article, measured in characters
\paragraph{prompt\_type} Type of prompt used (left-wing, unbiased, right-wing)
\paragraph{prompt\_category} Category of the prompt (instruction, completion)
\paragraph{length\_generated} Length of the generated text, measured in characters
\paragraph{repetition\_penalty} Indicates whether a repetition penalty was applied (yes/no)
\paragraph{generation} Text of the generated article
\paragraph{Self-BLEU\_generations} Self-BLEU score for evaluating the repetitiveness of the generated text
\paragraph{POLITICS\_human} Political bias of human-written article defined by Equation \ref{val} using the POLITICS model.
\paragraph{POLITICS\_generations} Political bias of generated article defined by Equation \ref{val} using the POLITICS model.
\paragraph{POLITICS\_shift} Political bias shift defined by Equation \ref{delta} using POLITICS\_human and POLITICS\_generations.
\paragraph{RoBERTa\_human} Political bias of human-written article defined by Equation \ref{val} using the RoBERTa model.
\paragraph{RoBERTa\_generations} Political bias of generated article defined by Equation \ref{val} using the RoBERTa model.
\paragraph{RoBERTa\_shift} Political bias shift defined by Equation \ref{delta} using RoBERTa\_human and RoBERTa\_generations.
\paragraph{PB\_supervised\_shift} Displayed political bias obtained by taking the average of RoBERTa\_shift and POLITICS\_shift.
\paragraph{PB\_Mistral\_Chat, PB\_Gemma\_Chat, PB\_Mistral, PB\_Chatgpt, PB\_Llama\_Chat} Text obtained from using the classification prompts from Table \ref{tab:CARP} with designated LLMs on a generated article
\paragraph{PB\_Mistral\_Chat\_Human, PB\_Gemma\_Chat\_Human, PB\_Mistral\_Human, PB\_Chatgpt\_Human, PB\_Llama\_Chat\_Human} 
Text obtained from using the classification prompts from Table \ref{tab:CARP} with designated LLMs on a human-written text
\paragraph{PB\_Mistral\_Chat\_Label, PB\_Gemma\_Chat\_Label, PB\_Mistral\_Label, PB\_Chatgpt\_Label, PB\_Llama\_Chat\_Label} 
Political bias value of the generated article, extracted from the output of the designated classification LLM
\paragraph{PB\_Mistral\_Chat\_Human\_Label, PB\_Gemma\_Chat\_Human\_Label, PB\_Mistral\_Human\_Label, PB\_Chatgpt\_Human\_Label, PB\_Llama\_Chat\_Human\_Label} 
Political bias value of the human-written article, extracted from the output of the designated classification LLM
\paragraph{Gemma\_Chat\_PB\_Shift, Mistral\_Chat\_PB\_Shift, Llama\_Chat\_PB\_Shift, Mistral\_PB\_Shift, Chatgpt\_PB\_Shift} Political bias shift for each designated LLM, calculated according to Equation \ref{delta}. This metric uses evaluations of both human-written and generated texts to determine the shift in political bias for each LLM
\paragraph{PB\_shift} Displayed political bias, calculated by averaging the values from Gemma\_Chat\_PB\_Shift, Mistral\_Chat\_PB\_Shift, Llama\_Chat\_PB\_Shift, Mistral\_PB\_Shift, and Chatgpt\_PB\_Shift
\\\\
Figures \ref{fig:WC_groups}, \ref{fig:num_of_sources}, \ref{fig:2hist}, \ref{fig:humanbleu}, and \ref{fig:bleu} then illustrate further properties of the dataset.


\clearpage
\newpage

\section{Further Political Bias Assessment Figures}
\begin{figure*}[h]
    \centering
    \includegraphics[width=0.85\textwidth]{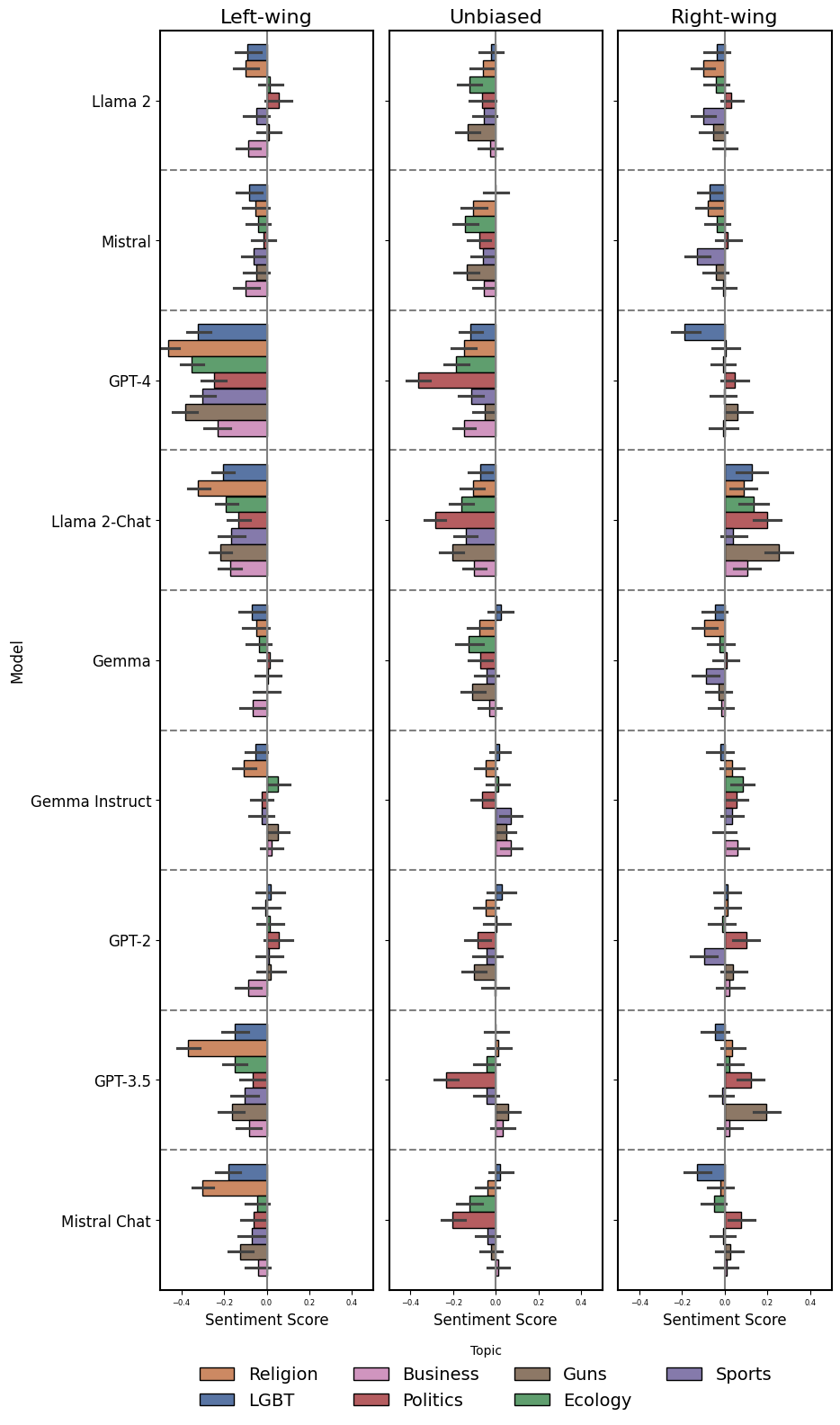}
    \caption{Political shift per model, prompt type, and news category (as assessed by supervised models)}
    \label{fig:PBS_large}
\end{figure*}

Due to our dataset being divided into categories, we are able to examine the political bias at this level as well. Figure \ref{fig:PBS_large} illustrates how varying prompt types result in different behaviours across news categories.

For left-wing prompts, religion and LGBT news appear to be most affected, with religion in particular exhibiting a pronounced shift to the left. Conversely, business and politics news seem to remain almost unchanged, showing only minimal shifts.

This behaviour is completely reversed for unbiased news prompts, where the politics category appears to be by far the most left-leaning. This almost suggests that LLMs tend to restrain themselves when discussing politics while being incentivised to display bias, but do not exhibit the same restraint when the news generation does not mention bias.

With right-wing prompt types, the topic of guns appears to have the greatest tendency to lean to the right. However, the bias is much weaker than that observed with left-wing prompt types.

\begin{figure*}[h]
    \centering
    \includegraphics[width=0.85\textwidth]{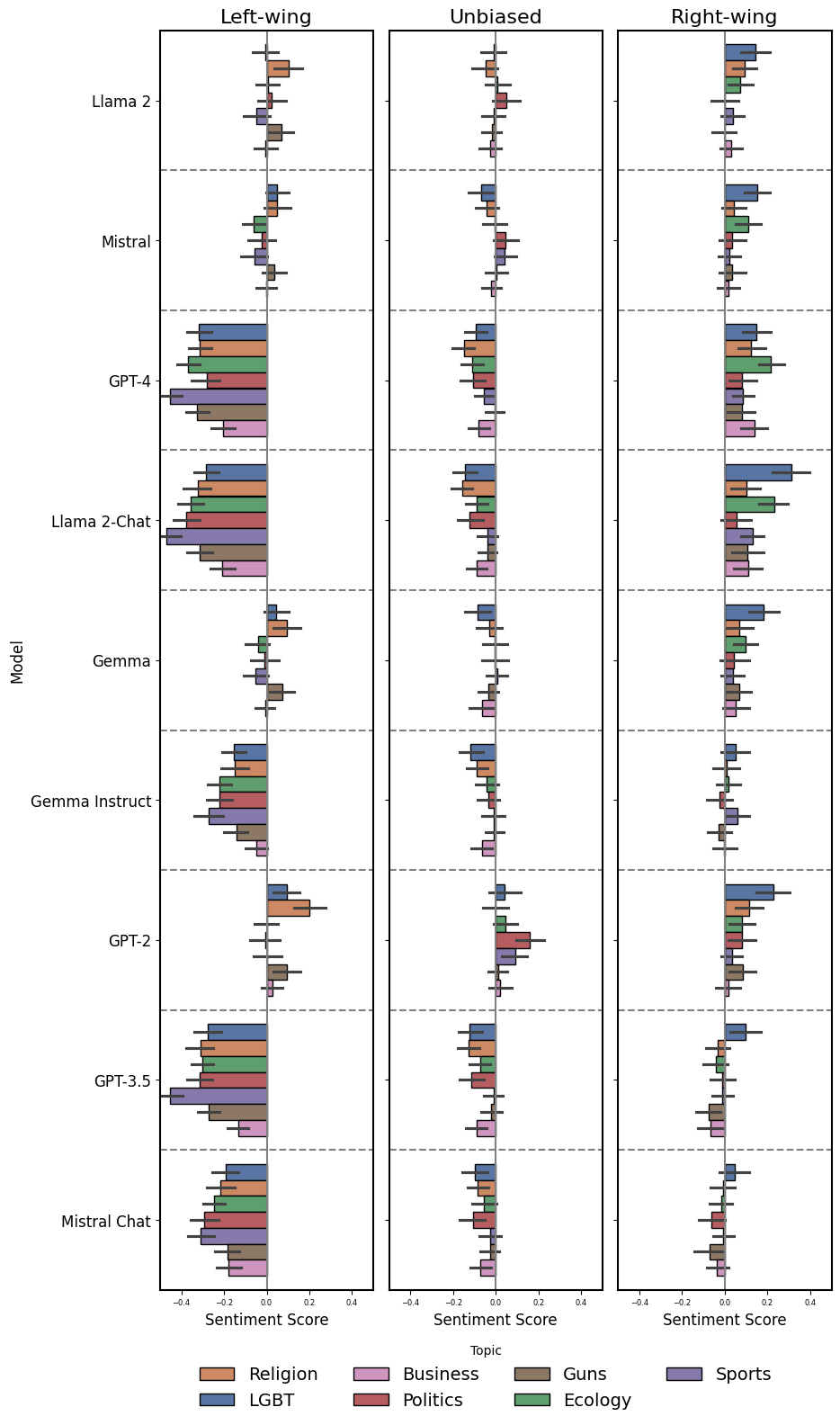}
    \caption{Political shift per model, prompt type, and news category (as assessed by LLMs)}
    \label{fig:PBS_large_LLM}
\end{figure*}

For the Figure \ref{fig:PBS_large_LLM}, which presents data from LLM classification, the trends appear to be much less distinct. The two most significant trends in this assessment are that the LGBT topic in right-wing prompts appears to exhibit a large rightward shift, and that the sports topic in left-wing news displays the strongest left-wing bias. Neither of these trends appears to be as pronounced when using the supervised models. This suggests that each classification model has a partially different perception of political bias and what constitutes left and right. 

\section{Manual Review Process}
The news generation, translation, and classification processes were all manually reviewed. This review involved authors examining 10 articles for each LLM used in generation. During this process, we looked for faults, degenerative outputs, and any offensive content. If faults were found, we altered the prompt and conducted another review, assessing both the new randomly selected set of 10 articles and the instances from the previous round where problems were identified. One issue found through this process was the repetition of the summary in the article generation, which we were able to mitigate.

\section{Scientific Artifacts}
In the course of this work, we employed many open-source scientific artefacts. These included HuggingFace Transformers \citep{wolf-etal-2020-transformers}, NumPy \citep{harris2020array}, Pandas \citep{pandas}, NLTK \citep{bird2009natural}, and PyTorch \citep{paszke2019pytorch}. Furthermore, we utilised multiple datasets, which are open-sourced for research purposes \citep{grusky-etal-2018-newsroom, baly-etal-2020-detect, TerAkopyan2021}. Consequently, we commit to making our work available, thereby opening opportunities for future follow-up and reproduction efforts.

\end{document}